\documentclass[journal]{IEEEtran}
\usepackage{blindtext}
\usepackage{graphicx}
\usepackage{amsmath}
\usepackage{algorithm}
\usepackage{algorithmic}

\DeclareMathOperator*{\argmax}{argmax}

\IEEEoverridecommandlockouts
\IEEEpubid{\makebox[\columnwidth]{978-1-7281-6286-7/20/\$31.00 \copyright2020 IEEE \hfill} \hspace{\columnsep}\makebox[\columnwidth]{ }}

\begin{document}
\title{Proactive Network Maintenance using Fast, Accurate Anomaly Localization and Classification on 1-D Data Series}

\author{Jingjie Zhu, Karthik Sundaresan, Jason Rupe\\~\IEEEmembership{CableLabs, Louisville, CO, U.S.A\\j.zhu@cablelabs.com, k.sundaresan@cablelabs.com, j.rupe@cablelabs.com}
}

\maketitle
\IEEEpubidadjcol

\begin{abstract}
Proactive network maintenance (PNM) is the concept of using data from a network to identify and locate network faults, many or all of which could worsen to become service failures. The separation between the network fault and the service failure affords early detection of problems in the network to allow PNM to take place. Consequently, PNM is a form of prognostics and health management (PHM).

The problem of localizing and classifying anomalies on 1-dimensional data series has been under research for years. We introduce a new algorithm that leverages Deep Convolutional Neural Networks to efficiently and accurately detect anomalies and events on data series, and it reaches 97.82\% mean average precision (mAP) in our evaluation.
\end{abstract}


\IEEEpeerreviewmaketitle

\section{Introduction}
PNM is a concept created around 10 years ago in the cable industry to help manage operations costs as service providers turned more attention toward bi-directional services. These data services needed resiliency mechanisms, which in turn provided opportunities for PNM. Receive modulation error ratio (RxMER) is a key data element for finding problems in the network. It is a measurement of the signal to noise on a per sub-carrier frequency basis. Each sub-carrier frequency carries a calibrated pilot signal periodically, which provides an opportunity to measure a calibrated signal level and calculate a signal to noise ratio. The result is a 1-dimensional data series of measurements over a span of frequencies. But for most cable operators today, translation of these data into a decision about whether there exists an impairment in the signal or not, how severe, and where it might be located, has been left to experts to decipher, and at times disagree. By introducing machine learning techniques to the problem, the industry believes that this step in the process can be automated and improved.

There are many challenges on detecting patterns from series of data. For instance, in anomaly detection on time series, traditional methods like auto-regressive integrated moving average (ARIMA) and sliding window median, are useful for detecting sudden changes in the data series. But they require well defined thresholds and specifically designed window sizes. These algorithms also cannot recognize the actual patterns of the anomalies and produce meaningful labels. Newer methods like long short-term memory (LSTM) can also be used for detecting patterns or anomalies on data series. However, compared to convolutional neural network (CNN), LSTM normally has more parameters and takes more time and resources to train, and it also has a tendency to overfit the training dataset. One of our previous research paper\cite{access_network_scte} uses compound sliding window median algorithm for region proposal generation and a fully-connected neural network for classification. However, the previous research has such a shortcoming that the region proposal module of the model is based on smoothing algorithms and is not able to accurately detect anomalies. In addition, the input of its classification neural network has data padding on each proposed anomaly, which significantly increases the amount of data that are being processed and results in poor inference performance.

Literature in object detection has been showing significant progress in recent years. Region proposal-based object detection algorithms such as faster R-CNN\cite{faster_rcnn} and regression based single shot algorithms such as YOLO\cite{yolov2}\cite{yolov3}, SSD\cite{ssd}, and Retinanet\cite{retinanet} are showing great performance in object detection problems. Given that image data are far more complex than data series, we adopt the architecture of YOLOv3\cite{yolov3} and develop a single-shot anomaly detector which provides superb efficiency and accuracy for recognizing anomalies and events on data series. This novel algorithm uses a Fully-Convolutional Neural Network and custom feature aggregation layers and prediction layers to perform localization and classification in a single step, which significantly improves the localization accuracy and classification performance compared to the previous research\cite{access_network_scte}. In use cases such as detecting impairments using cable modems' Orthogonal Frequency-Division Multiplexing (OFDM) and Orthogonal Frequency-Division Multiple Access (OFDMA) RxMER data, we have demonstrated the effectiveness of our algorithm.

\section{Background}
\subsection{DOCSIS\textregistered{} Specification Context}
The DOCSIS protocol enables two-way radio-frequency (RF) communication over coax, a technology providing much of the access network for entertainment programming and internet access for the world, not to mention a large amount of networking for businesses.

It relies on analog RF transmission of data, and uses a mix of methods including single carrier quadrature amplitude modulation, OFDM/A (3.1), SC-QAM (3.0), etc., depending on the version and implementation. 

For a long time, it has included several resiliency mechanisms including Forward Error Correction (FEC), adjustable profiles with differing data rates per frequency carrier, echo cancellation, and equalization. These resiliency mechanisms adjust service to account for plant problems that create impairments to the RF transmission signal. When these adjustments are made, it indicates in many cases an imperfect transmission medium. When this imperfection goes beyond the designed and installed quality level, it indicates network (plant) damage or degradation. This mechanism therefore can be monitored to indicate changes or problems in the network (plant). The operator can know of a problem before the customer is impacted. When a problem is indicated, a proactive repair opportunity may be created. PNM is the industry term for addressing plant problems before service is impacted. 

\subsection{Proactive Network Maintenance Description}
PNM was envisioned in the cable industry more than a decade ago, when DOCSIS networks were relatively new, and resiliency mechanisms were creating the opportunity\cite{larry_2015}. Several measures were identified based on then-understood failure modes and risks associated with deployed technology. At the time, for example, analog optics were common for feeding the coax plant, so in some systems optical clipping was more common than it is today with digital optical systems. But the coax plant remains and will for some time, bringing with it several failure modes that result in RF transmission issues in certain frequencies, and impact on subsequent layers of the communication connection. 

General components and their failure modes in the outside plant that can appear as impairments in the RF signal include but are not limited to the following:
\begin{itemize}
\item Hard line or drop cable – shield, sheath, conductor, insulator
\item Connector – center conductor, shield, thread
\item Tap – cover, connector, electronics
\item Splitter, combiner, coupler – cover, connector, electronics
\item Filter – cover, connector, electronics
\item Splice – wrap, fill, shield connection, conductor connection, strap 
\item Amplifier – cover, connector, electronics
\item Power Supply – cover, connector, electronics
\item Anchoring – various types such as straps, guy wires, and more. 
\item Node – cover, connector, electronics
\end{itemize}

General failure modes of the cable plant which impact RF signal include the following
\begin{itemize}
\item Passive – loose or misaligned connection, corroded, poor installation, damage, degradation, wet, cracked, loose and moving, incorrect part. 
\item Active – poorly made, poorly installed, electrostatic discharge (ESD), lightning, ground fault, degradation or wear out, damaged, incorrect part. 
\end{itemize}

Measurements identified for proactive network maintenance in the specification include the following, though some of these measurements are more general and not specifically identified for PNM:
\begin{itemize}
\item Downstream spectrum capture
\item Downstream symbol capture
\item Downstream channel estimation coefficient
\item Downstream constellation display
\item Downstream receive modulation error ratio (RxMER) per sub-carrier
\item Downstream forward error correction (FEC) summary
\item Downstream required quadrature amplitude modulation (QAM) MER 
\item Downstream histogram 
\item Downstream modulation profile 
\item Upstream pre-equalization 
\item Upstream spectrum capture
\item Upstream RxMER per sub-carrier
\item Upstream FEC summary
\item Power level
\end{itemize}

For the version of the anomaly detector described in this paper, we will focus on downstream RxMER per sub-carrier as the measure used. While not a complete solution for identifying and removing RF impairments in coax networks, it is an important measure and can address a very significant part of PNM. RxMER per subcarrier is a measure of the signal to noise of each downstream frequency used to carry data to the cable modems. The cable modem can report these data because it monitors the frequencies over which it receives data, and those frequencies occasionally carry calibrated pilot signals. Those signals are known, so they can be compared against a received signal to determine a signal to noise ratio. Ideal signals are mostly flat across all subcarriers, with small amounts of variation acceptable. More detail about this measure can be found in\cite{docsis31phy}\cite{docsis31ossi}\cite{larryscte}\cite{ref11}\cite{proops_report}. Many impairments show up in RxMER per sub-carrier data, visually, statistically, and otherwise\cite{ref11}.

Impairments that appear in the spectrum (and RxMER per sub-carrier data) are classified in the cable industry as follows
\begin{itemize}
\item Standing waves – when an RF signal transmits though coax and encounters an impedance change, the signal loses strength in the forward direction, and an echo returns in the opposite direction. If there are two impedance mismatches in the line, then an echo tunnel can form, causing repeating echo signals. When the impedance mismatches are stable, the echo tunnel appears in the RF signal as a standing wave, which impedes the signal in at least two ways: energy loss due to the impedance mismatches, and noise from the echo. Both of these factors impact the RF signal. Echo cancelers can take care of some of the impact at some points, but only if there is a signal with which to cancel. Power levels can adjust somewhat for fixed signal level effects, but not in all cases.
\item Resonant peaking – sometimes in frequency data we see certain frequencies with higher energy than others. We have not seen this in RxMER per sub-carrier data, and due to the nature of the measurement we don’t expect to encounter a resonant peaking, though the same pattern may appear in the data due to other frequencies experiencing impairment issues.
\item LTE or FM Ingress – when there are opens in cable plant, RF from the outside can get into the cable network. FM and LTE cellular signals (as do VHF and UHF bands) overlap the spectrum used in DOCSIS networks, so can appear as noise when the cable plant is open due to shield failures (shield integrity problems).
\item Suck-out – An ideal cable plant transmits evenly on all frequencies, but actual cable plant transmits better at lower frequencies than at high frequencies, so amplification is adjusted to make it behave more ideally. Typically, we see this as a slope in the power levels, when not adjusted for in transmission. But grounding issues in the cable plant can lead to some frequencies being highly attenuated when they should not be. These appear as a rapid drop in a band of frequencies, like the energy is being sucked out of the system at that frequency. The frequencies affected by the suck-out become unusable when severe enough, and at least must be compensated for otherwise. Because these problems can come and go over time and conditions, and often get worse over time, these are a perfect opportunity for PNM.
\item Roll-off – at the edge of a frequency band, frequencies at the edge can be attenuated more as you get closer to the edge. This is called a roll-off, and results in frequencies not being useful near that edge.
\item Filter – band filters in the plant can protect noise from getting into the system, but sometimes filters can attenuate frequencies on band edges unintentionally. Finding and removing these issues improves service.
\item Adjacency, second order distortion, and phase distortion are less common. Adjacency may appear as a step in the data over some sub-carriers. Second order distortion may appear as noise in RxMER per sub-carrier, and could potentially be captured by correlating with carrier signals whose energy appears in other frequencies, though detecting this impairment type requires correlating with the sent signal. Similarly, phase distortions require more than magnitude data (which RxMER per sub-carrier is) or potentially correlating with the complex I and Q values of the signal.
\end{itemize}

As access networks are physically a tree structure, an impairment has different impact on customers connected at different places of the plant. Taps include filters that protect customers from some issues. Recognizing RxMER is a measurement of the signal to noise that is received by the cable modem, and it is a measurement of the pilot signals only (which are of a known energy level), not all network problems will be discovered with RxMER. Further, due to the design of cable plant with diplex filters and taps, some impairments further downstream from a given cable modem may not impact that cable modem. These features help in localizing the problem sometimes.

The distinction between the network fault and service failure is a lot like the distinction between a software fault and software execution failure. In software, a fault may not lead to a failure in all cases, and may not be executed frequently or ever in an application. Likewise, a network fault may be hidden by resiliency mechanisms that come with DOCSIS networks and well-designed cable plant, or the impacted frequencies may not be used or used at lower data rates to compensate such that service is not impacted. Yet these impairments can grow to impact service as a failure, or service conditions can change such that the fault becomes a service failure.

\subsection{ProOps Environment}
We created the Proactive Operations\cite{proops_report}\cite{proops_scte} (ProOps) platform to enable automation of detection of PNM issues into operations processes that turn the data into action. 
This platform enables code modules, referred to as workers, that process data input to form statistics output. The data can be raw data, statistics, processed data, or otherwise. The output can be any statistic including a hard decision point. 
We direct but do not require a four layer architecture for organizing the workers in ProOps to fulfil one or more of the following layers
\begin{itemize}
\item Observe – collect data from the network elements (usually cable modems identified by MAC address) and process into soft decisions or statistics for further analysis.
\item Orient – analyze the data from the observe layer and decide what additional data are needed, or more frequent collection of some data, or a broader scope of the collection of those data such as over neighbors or over longer periods of time, for example.
\item Decide – analyze the data, statistics, and soft decisions from the first two observe and orient layers to turn found anomalies in the data into network impairments, through classification of anomalies, clustering the data from the network elements, and applying either network health or severity measurements to quantify the indicated problems.
\item Act – by organizing the found problems by severity or network health scores, we can allow the operator to select which found issues deserve attention, and what type of human intervention or further work is needed, be that a field technician or someone in a network operations center.
\end{itemize}

ProOps is depicted in Figure 1 below, with the four layers represented. Also shown are the configuration interface, control and scheduling functions, the work queue for workers to process, and the various types of data stores used. The results can be observed on a dashboard, which can include maps showing impairments colored by severity, or graphs of the network health or impact of impairments that rise to attention, for example.

ProOps has been built with several types of workers, and can be configured to use these in various ways. For the sake of this paper, we will limit the discussion to the anomaly detector which will process RxMER data from cable modems and identify anomaly types at the first layer. The subsequent layer will collect additional information to assist in quantifying the severity of the found anomalies. Then the next layer will simply match impairment patterns as a simple clustering, and calculate a severity. The final layer presents a sorted list of the identified impairments to be addressed from most severe to least. At the point of this writing, only the clustering mechanism has not yet been built; we anticipate using a k-means method or like matching algorithm. The focus of this paper is the anomaly detection worker which is very efficient and effective at this particular problem.

\begin{figure}[h!]
  \centering
  \includegraphics[width=0.45\textwidth]{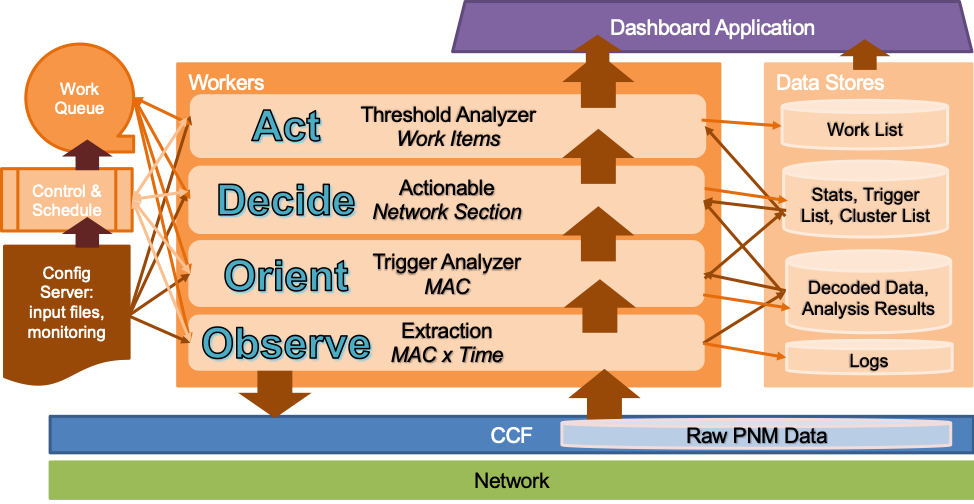}
  \caption{Proactive Operations Platform}
\end{figure}

\section{Model Architecture}

\begin{figure*}[t]
  \centering
  \includegraphics[width=\linewidth]{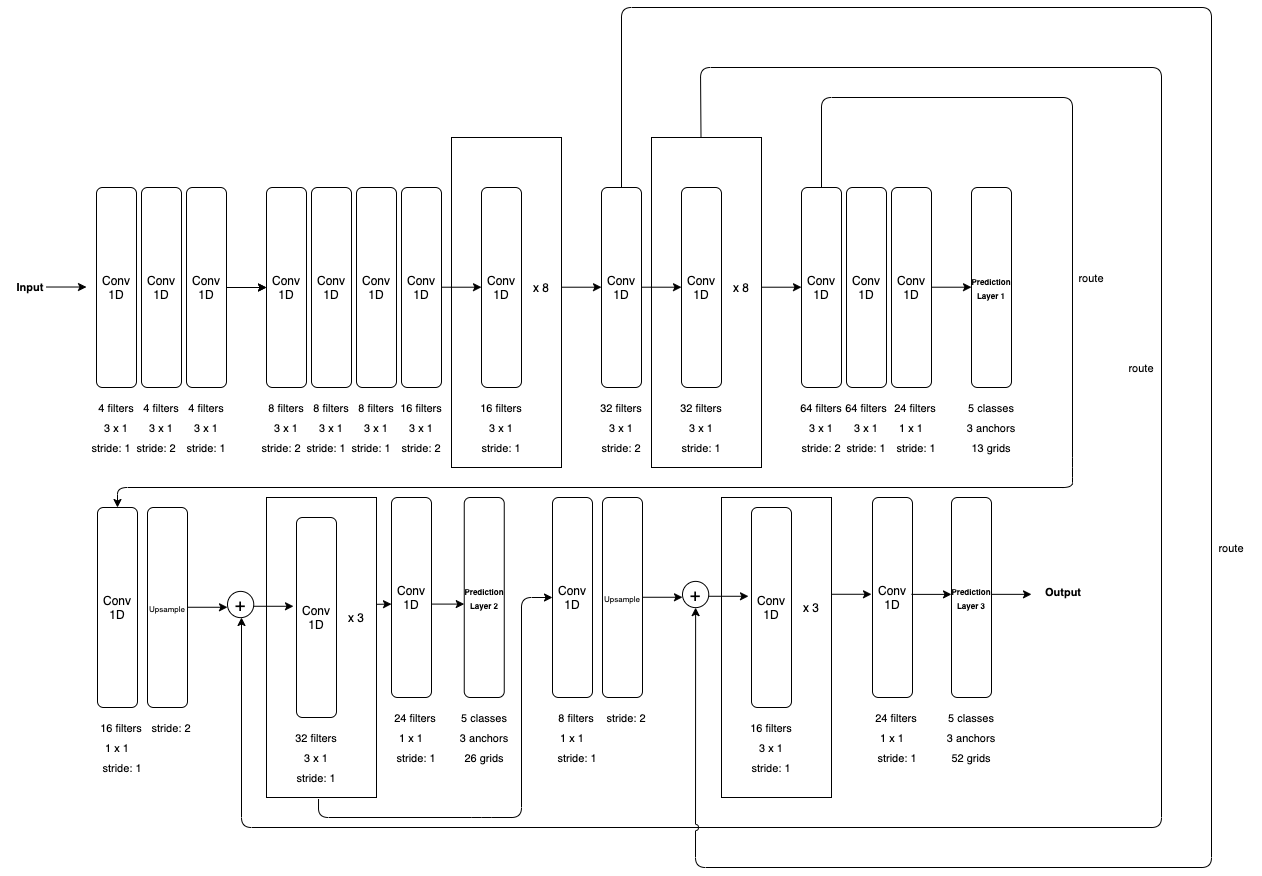}
  \caption{Model Architecture}
  \label{nn_arch}
\end{figure*}

The architecture diagram Fig.\ref{nn_arch} shows the high level architecture of our model. The one dimensional convolutional layers, route layers, up-sampling layers, and 3 prediction layers are included in the diagram. The shortcut layers, activation layers, and batch normalization layers are omitted in the diagram for simplicity.

\subsection{Backbone Network}
The backbone network consists of 1-D convolutional layers with 1-D filters for feature extraction and shortcut layers for feature aggregation. The first 3 convolutional layers only have 4 filters with kernel size 3 and the maximum number of filters in the following layers is 64. This is because 1-dimensional data have much simpler features such that a smaller network can reach high mAP while featuring great performance and low resource consumption. Zero padding is used on both sides of the sample before convolution when the kernel size is greater than 1. This allows the network to maintain the consistency and relevance of the feature map sizes.

The shortcut layers perform addition on the feature maps from the previous layer and the layer it points to. It has multiple benefits such as smoothing the optimization during training and preventing the model from being attracted by spurious local optima as we learned in the previous study\cite{shortcutstudy}.

\subsection{Route Layers}
The Route layers are used for feature concatenation. When a route layer only points to one previous layer, the feature map output from the previous layer is forwarded as the output of the route layer. When a route layer points to multiple previous layers, their feature maps are concatenated and forwarded as the output of the route layer. In our network, features from lower level layers that contain more original information are concatenated with outputs from higher level layers to provide more information to the prediction layer to improve mAP.

\subsection{Up-sampling Layers}
The up-sampling layers are used to increase the size of the feature map from the previous prediction layer (the layer that predicts at larger grids) to match the size of the forwarded feature map from lower level layers. In our network we use a nearest neighbor interpolation algorithm for the up-sampling process.

\subsection{Activation Layers}
Activation functions are implemented as layers in our network. All 1-dimensional convolutional layers are followed with Leaky Rectified Linear Unit (Leaky ReLU) activation layers except the 3 convolutional layers before the 3 prediction layers. The Leaky ReLU activation is calculated as
\begin{equation}
\text{leaky} \equiv
                \begin{cases}
                x,           & \text{if $x \geq 0$}\\
                \alpha x,    & \text{if $x < 0$}
                \end{cases},
\end{equation}
where $\alpha$ is 0.1 in our network.

\subsection{Batch Normalization Layers}
We use batch normalization layers\cite{batchnorm} in between 1-dimensional convolutional layers to normalize the input layer by adjusting and scaling the activations. Batch normalization improves the training speed, stability, and mAP of our network.

\subsection{Prediction Anchors}
Instead of predicting the start, end, and the center of anomalies directly, we use anchors which are a set of hand-picked priors\cite{faster_rcnn} that the prediction layers use as references. This simplifies the localization problem and makes it easier for the network to learn\cite{yolov2}. With this, the center prediction becomes
\begin{equation}
p'_{center} = \sigma(p_{center}) + g_{center},
\end{equation}
where $p'_{center}$ is the final value of the predicted anomaly center, $\sigma(p_{center})$ is the network predicted value $p_{center}$ calculated with sigmoid function
\begin{equation}
\sigma(x) =  \frac{\mathrm{1} }{\mathrm{1} + e^{-x}}, 
\end{equation}
and $g_{center}$ is the center of the grid that is performing prediction where a grid is defined as a prediction unit that takes input data from one feature map channel and handles all anomalies which have their center points inside of it. The width prediction of the anomaly becomes
\begin{equation}
p'_{width} = e^{p_{width}} \cdot \beta_{width},
\end{equation}
where $p'_{width}$ is the final value of the predicted anomaly width, $e^{p_{width}}$ is the network predicted value $p_{width}$ calculated with exponential function, and $\beta_{width}$ is the width of the current anchor.

\subsection{Anchor Calculation}
We use both K-Means and kernel density estimation (KDE) algorithm to calculate proper anchor sizes. This allows the sizes of the prediction anchors to fit the distribution of anomaly width values in datasets as much as possible, and makes the problem easier for our network to learn. In our experiments with cable modems' downstream RxMER data, the width value of anomalies ranges from 2 to 416 (with input size 416) as shown in Table \ref{tab:anchors}.
\begin{table}[h!]
\renewcommand{\arraystretch}{1.3}
\caption{Anchor sizes at each prediction layer}
\label{tab:anchors}
\centering
\begin{tabular}{|p{3cm}|p{2.2cm}|p{2.2cm}|}
    \hline
    Prediction Layer & Anchor Sizes\\
    \hline
    
    \hline

    Layer 1 (13 prediction grids) & 155, 234, 416\\
    \hline

    Layer 2 (26 prediction grids) & 43, 73, 109\\
    \hline
    
    Layer 3 (52 prediction grids) & 2, 8, 23\\
    \hline
\end{tabular}
\end{table}

\subsection{Prediction Layers}
With specifically designed convolutional layers as the backbone, at the first prediction layer, the size of the input feature map is transformed to
\begin{equation}
S_{feature} = 1 \times \Big[(3 + n_{classes}) \times n_{anchors}\Big] \times \Big( \frac{S_{input}}{2^{n_{\gamma}}} \Big),
\end{equation}
where $S_{feature}$ is the total number of values in the feature map, $S_{input}$ is the original input size, $n_{classes}$ is the number of classes in the training and testing dataset, $n_{anchors}$ is the number of anchors that are being used by the current prediction layer, and $n_{\gamma}$ is the number of convolutional layers in the backbone network with stride 2.

The number of grids doubles when it becomes the second prediction layer for detecting mid-sized anomalies, and it doubles again when it comes to the third prediction layer for small anomalies. For instance, when using $1 \times 416$ as the input sample size and there are 5 anomaly classes to predict and 3 anchors at each prediction layer, the size of the input feature map from a single sample at the first prediction layer is $1 \times 24 \times 13$. The size becomes $1 \times 24 \times 26$ at the second prediction layer and $1 \times 24 \times 52$ at the third prediction layer.

Looking further into the anchor prediction vectors from the above example, each of the three anchors predicts a value $p_{center}$ for the center of the anomaly, a value $p_{width}$ for the width of the anomaly, a confidence score $p_{conf}$ for the objectness of the anomaly, and 5 probabilities for each anomaly class. The predictions are initially labeled using softmax function on the class probabilities (if it is multi-label classification, classes with top $n$ probabilities are selected) and then filtered by a confidence threshold, which is $0.5$ in our network. Non-maximum suppression (NMS) is then used (only in inference process) to reduce the number of localization proposals on each anomaly. In our model, we apply NMS to each class predictions instead of all predictions. The final output is a list of predictions each consists of a class label(s), an anomaly confidence score, the center location of the anomaly (proportional), and the width of the anomaly (proportional).

\section{Training}

\subsection{Datasets}
In our experiments, we use downstream receive modulation order ratio (RxMER) per sub-carrier data captures from cable modems' OFDM channels in DOCSIS 3.1\cite{docsis31phy} networks to test our algorithms and implementation. DOCSIS 3.1 cable modems are capable of capturing PHY layer data like downstream RxMER per sub-carrier based on requirements defined in DOCSIS 3.1 CM OSSI specification\cite{docsis31ossi}, which makes it a promising use case for real-world demonstration.

\begin{figure}[h!]
  \centering
  \includegraphics[width=0.45\textwidth]{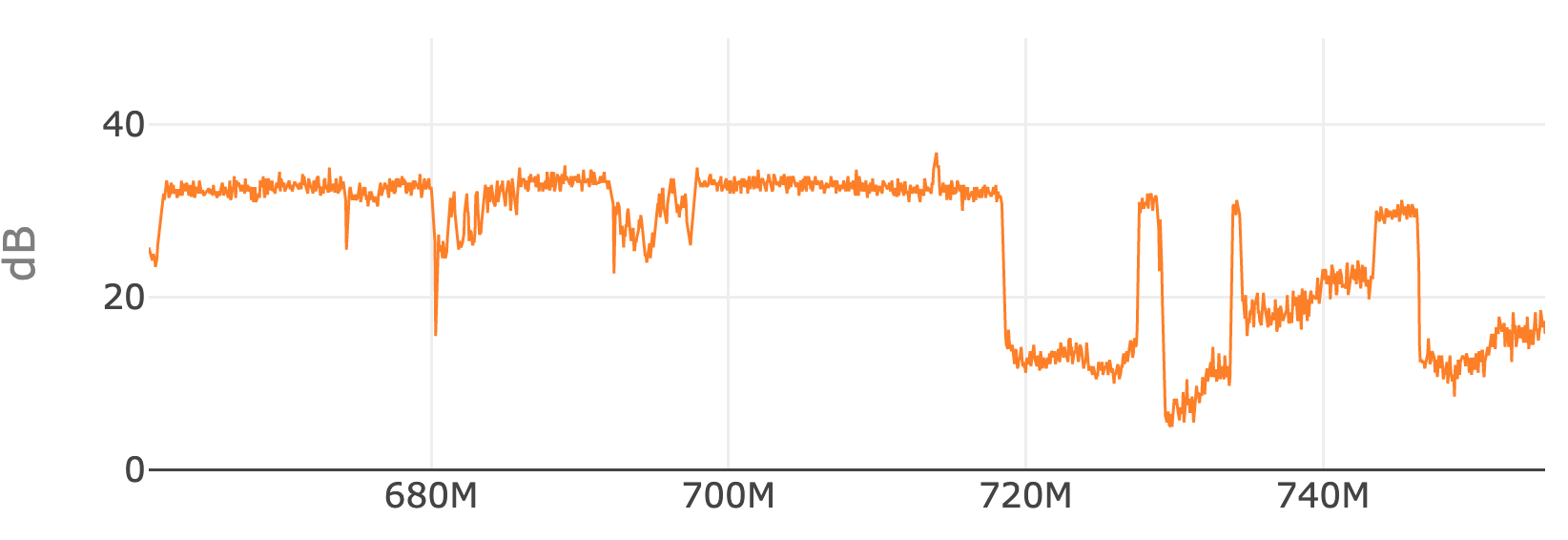}
  \caption{Downstream OFDM RxMER capture with multiple LTE ingress points and spikes}
  \label{multiple_lte}
\end{figure}

\begin{figure}[h!]
  \centering
  \includegraphics[width=0.45\textwidth]{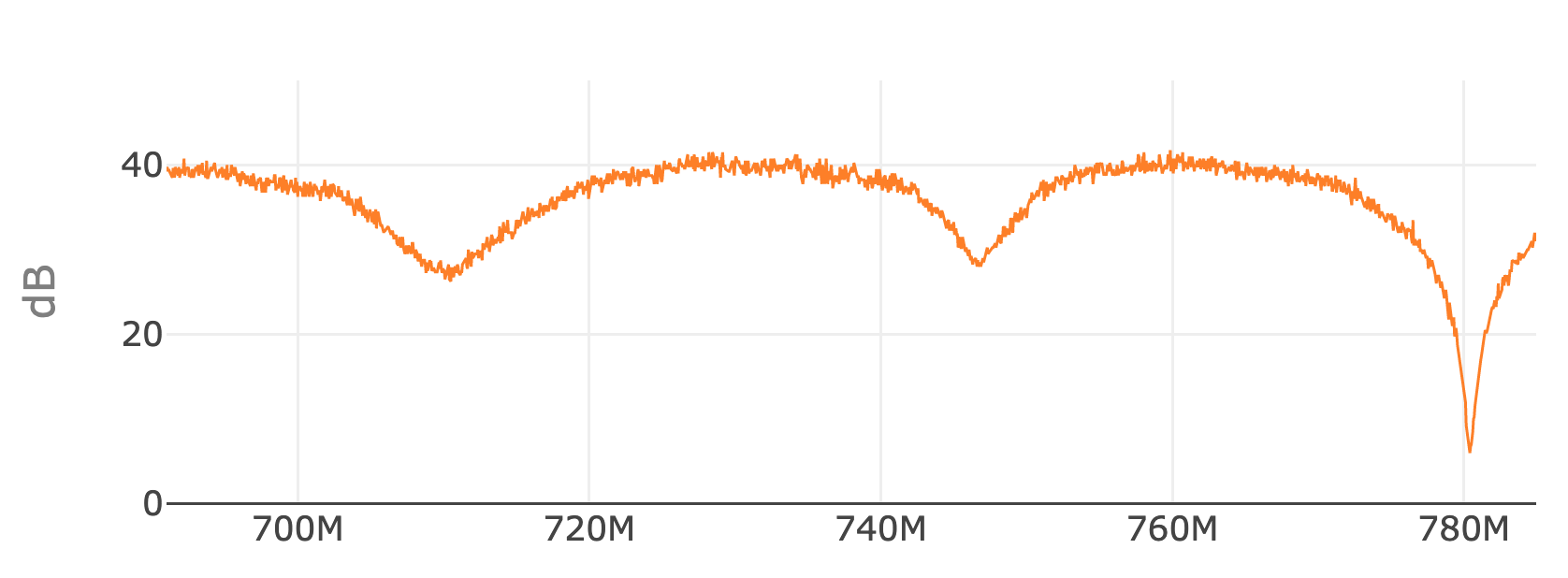}
  \caption{Downstream OFDM RxMER capture with multiple suck-outs}
  \label{multiple_suck_outs}
\end{figure}

\begin{figure}[h!]
  \centering
  \includegraphics[width=0.45\textwidth]{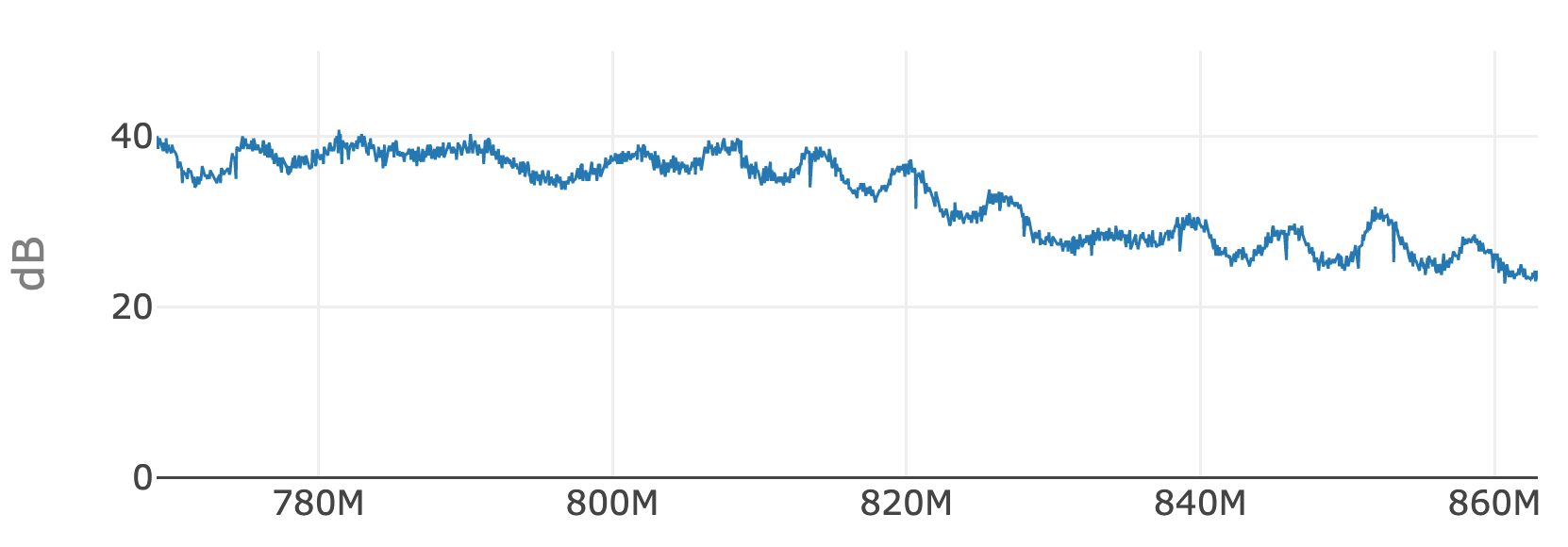}
  \caption{Downstream OFDM RxMER capture with a wave}
  \label{wave_fig}
\end{figure}

\begin{figure}[h!]
  \centering
  \includegraphics[width=0.45\textwidth]{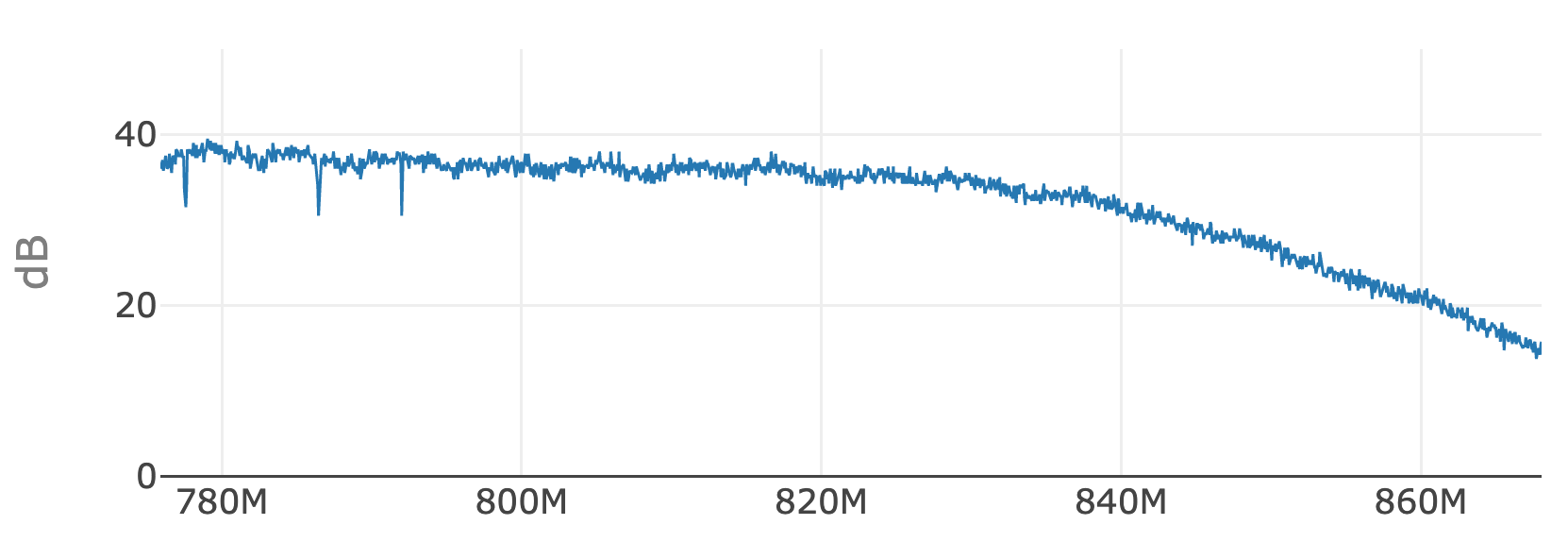}
  \caption{Downstream OFDM RxMER capture with a right roll-off and 3 spikes}
  \label{roll_off}
\end{figure}

The training dataset contains 45000 RxMER samples and the testing dataset contains 26000 RxMER samples. Each sample has 1800 to 2000 sub-carrier MER values that are in range $[0, 63.75]$. There are 5 classes of PHY layer impairments that are labeled in the datasets: LTE ingress noise, wave, roll-off, suck-out, and spike. The LTE ingress noise is normally identified as LTE signals that are adding interference to the OFDM signal through leakage points on the plant or defected shielding on the modem as shown in Fig.\ref{multiple_lte}; wave is identified by wave shaped impairments (as shown in Fig.\ref{wave_fig}) that affect the whole spectrum or part of the spectrum and is normally caused by echos or impedance mismatch; roll-off is identified as rolling off RxMER values on either side of the OFDM spectrum as shown in Fig.\ref{roll_off}; suck-out is identified as dips with a sharp corner as shown in Fig.\ref{multiple_suck_outs}, and is normally caused by issues on amplifiers; spike is identified as sharp dips as shown in Fig.\ref{multiple_lte} and Fig.\ref{roll_off}, and are normally less than 3 sub-carriers wide. Each impairment on the sample is labeled with the center of the impairment $x$ (proportional), the width of the impairment $w$ (proportional), and the class $c$ of the impairment. Each RxMER sample can have multiple different types of impairments located at different frequency ranges. Example plots of RxMER per sub-carrier data captured from cable modems are shown in Fig.\ref{multiple_lte} through Fig.\ref{roll_off}, showing various impairment types.

\subsection{Synthetic Data Generation}
To reduce labor work and increase the number of training samples, we randomly generate synthetic training data based on human-observed samples and anomalies. For instance, in our datasets, we use randomly generated noisy base samples as the starting point: spikes are generated randomly on different frequencies with different MER values; suck-outs are generated using two randomized non-linear functions; waves are generated from different sine waves; roll-offs are generated with one non-linear function. All anomalies are added to different locations on the sample.

The synthetic data are used as an addition to the human labeled datasets for both training and testing. It significantly increases the amount of data that joins training, which helps the model to generalize better. And it adds more samples that are unseen by the model for validation.

\subsection{Data Pre-processing with Binning Down-sampling}
While all input values are normalized to range $[0, 1]$, the binning minimum algorithm (algorithm 1) is used for preserving low values in the original data series while reducing the input sample size. The algorithm first performs up-sampling (nearest-neighbor algorithm) on the original sample to make it become the nearest dividable size $D_{nearest}$ by the target sample size, then calculates binning minimum values from the up-sampled input.

\begin{algorithm}
\caption{Binning Minimum Down-Sampling}
\begin{algorithmic}
\REQUIRE $S_{size} > 0, T > 0, S_{size} \geq T, L = []$
\STATE $m = S_{size} \mod T$
\STATE $n = (int)S_{size} / T$
\IF{$m \neq 0$}
\STATE $n = n + 1$
\ENDIF
\STATE $Interporlate(S, n \times T)$
\STATE assert $S_{size} \equiv n \times T$
\STATE $i = 0$
\WHILE{$i < T$}
\STATE $B_{min} = \min(S[i \to i + n - 1])$
\STATE $L \leftarrow$ append $B_{min}$
\STATE $i = i + n$
\ENDWHILE
\RETURN $L$
\end{algorithmic}
\end{algorithm}

\subsection{Data Augmentation}
During training, multiple data augmentation techniques are used to improve training results, improve the model's generalization ability, and increase the number of training samples. Each of the data augmentation techniques has a probability of being selected for each input sample.

\subsubsection{Scale Shifting}
The scale shifting process allows the network to learn samples with different noise fluctuation scales. The mean value of the sample is calculated at first as
\begin{equation}
\mu_{sample} = \frac{1}{n} \sum_{i=1}^{n}v_{i},
\end{equation}
where $\mu_{sample}$ is the mean value of the sample, $n$ is the number of values in the sample array, and $v_{i}$ is the $i_{th}$ value.

Then for each original value
\begin{equation}
v_{i} = (v_{i} - \mu_{sample}) \cdot R_{scale} + v_{i},
\end{equation}
where $R_{scale}$ is a randomly generated factor in range of $[0.3, 3]$.

\subsubsection{Flipping}
Flipping allows the network to learn anomalies at different locations and doubles the total number of samples for training. The input sample is randomly flipped (left-right), and the labels are converted if the sample is flipped.

\subsubsection{Value Floor Shifting}
By modifying the floor of the input sample, the network can adapt to anomalies at different values levels. For each value $v_{i}$ in the input sample
\begin{equation}
v_{i} = v_{i} + R_{level},
\end{equation}
where $R_{level}$ is a randomly calculated value in range of $[-0.2, 0.2]$.

\subsubsection{Noise Injection}
Random noise (not enough to become anomalies) is added to each value in the input sample to allow the network to adapt to noisy inputs and multiplies the number of different samples for training,
\begin{equation}
v_{i} = v_{i} + R_{noise},
\end{equation}
where $R_{noise}$ is a randomly generated value in range $[-0.002, 0.002]$ for each value in the sample.

\subsubsection{Smoothing}
A Savitzky-Golay filter is used to reduce the amount of high frequency noise in the input sample. This allows the network to adapt to smooth inputs and again multiplies the number of different samples for training. Different filter window lengths (3, 5, 7) are randomly selected during training for different input samples.

\subsubsection{Cut and Paste}
We randomly move (cut and paste) anomaly objects on the x-axis of the input sample to improve the training efficiency on every prediction grid. This increased the mAP on detecting small and dense anomalies on cable modems' OFDM receive modulation error ratio per sub-carrier data by $4\%$.

\subsection{Optimizer and Hyper Parameters}
Stochastic gradient descent (SGD) is used as the optimizer in our experiments with 6000 burn-in mini-batches, 0.9 momentum, $1e^{-3}$ learning rate, $5e^{-4}$ weight decay, and mini-batch size 32. During the burn-in time, the learning rate increases gradually until it reaches the target learning rate
\begin{equation}
\alpha_{current} = \min\{\alpha_{target} (\frac{N_{batches}}{N_{burn in}})^4, \alpha_{target}\}
\end{equation}
where $\alpha_{target}$ is the target learning rate, $\alpha_{current}$ is the learning rate of the current iteration, $N_{batches}$ is the number of trained mini-batches, and $N_{burn in}$ is the number of burn-in mini-batches.

\subsection{Loss Calculation}
Our loss function for training the model is designed with reference from YOLOv2\cite{yolov2}. To make it more efficient for our anomaly detection task and better balance loss calculation between small anomalies and large anomalies, we add scale weights in the mean square error (MSE) loss for localization loss calculation. This improves the model's ability to localize small anomalies more precisely.

The localization loss of the center prediction is calculated as
\begin{equation}
L_{1} = \sum_{i=0}^{N_{grid}}\sum_{j=0}^{N_{anchor}}C_{i} \Big\{[\gamma_{i}(x_{i} - \hat{x_{i}})]^2 + [\gamma_{i}(\sqrt{w_{i}} - \sqrt{\hat{w_{i}}})]^2\Big\},
\end{equation}
where $N_{grid}$ is the number of prediction grids at the current prediction layer, $N_{anchor}$ is the number of anchors used at the current prediction layer, $C_{i}$ is the ground truth confidence which value can be 0 or 1, $x_{i}$ is the ground truth center location (proportional) of the anomaly, $\hat{x_{i}}$ is the predicted center location (calculated with the anchor) of the anomaly, $w_{i}$ is the ground truth width (proportional) of the anomaly, $\hat{w_{i}}$ is the predicted width (calculated with the anchor) of the anomaly, and $\gamma_{i}$ is the scale weight of the anomaly
\begin{equation}
\gamma_{i} = 2 - w_{i},
\end{equation}
where $w_{i}$ is in range $(0, 1]$ and anomalies with smaller width have larger weights. The confidence loss is designed for the network to converge to a point that background data corresponds to confidence score 0, and anomaly data corresponds to confidence score 1. The confidence score is first calculated from the prediction layer's linear output with sigmoid function
\begin{equation}
\hat{C_{i}} = \sigma(\hat{C'_{i}}).
\end{equation}
The confidence loss calculation is based on binary cross-entropy (BCE) loss, but use different weights for anomalies and background in order to balance the network's recall and precision
\begin{equation}
L_{2} = \sum_{i=0}^{N_{grid}}\sum_{j=0}^{N_{anchor}}-1 \cdot \lambda^{conf}_{i}[C_{i}\log(\hat{C_{i}}) + (1 - C_{i})\log(1 - \hat{C_{i}})],
\end{equation}
where $C_{i}$ is the ground truth confidence score, $\hat{C_{i}}$ is the predicted confidence score, $\lambda^{conf}_{i}$ is the weight of the confidence loss which can be represented as
\begin{equation}
\lambda^{conf}_{i} \equiv
                \begin{cases}
                1,      & \text{if $C_{i} = 1$}\\
                0.5,    & \text{if $C_{i} = 0$}
                \end{cases},
\end{equation}

The classification loss is calculated based on BCE loss as
\begin{equation}
\begin{aligned}
L_{3} = {} & \sum_{i=0}^{N_{grid}}\sum_{c \in classes}-1 \cdot C_{i}[p_{i}(c)\log(\hat{p_{i}}(c)) \\ & + (1 - p_{i}(c))\log(1 - \hat{p_{i}}(c))],
\end{aligned}
\end{equation}

\begin{figure}[h!]
  \centering
  \includegraphics[width=0.30\textwidth]{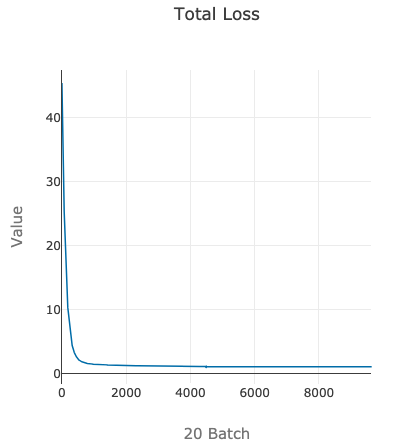}
  \caption{Model convergence (each unit is 20 mini-batches)}
  \label{fig_6}
\end{figure}

The total loss is calculated as
\begin{equation}
L_{total} = L_{1} + L_{2} + L_{3}.
\end{equation}
The convergence of the model is shown in Fig.\ref{fig_6}.

\subsection{Responsible Prediction Layer and Anchor Selection}
For each anomaly during training, only 1 prediction layer out of all 3 prediction layers is responsible for the prediction, and only 1 anchor out of all 3 anchors is responsible. The responsible prediction layer and anchor are selected by calculating which anchor has the best intersection over union (IoU) against the ground truth anomaly ignoring its center location
\begin{equation}
I_{anchor} = \argmax_{i \in anchors} \frac{\min(w, A_{i})}{\max(w, A_{i})}
\end{equation}
where $I_{anchor}$ is the responsible anchor's index, $w$ is the current anomaly's width, and $A_{i}$ is the $i_{th}$ anchor's width. Once the index of the responsible anchor is calculated, the responsible prediction layer is determined by where the responsible anchor is used. Because the loss of all the other anchors used by 3 prediction layers' is calculated as that these anchors see background with no anomaly, the ground truth confidence scores for these anchors are 0. The predictions from anchors which have IoU with the ground truth that are higher than the ignoring threshold $T_{ignore}$, which is 0.7 in our experiments, do not join the loss calculation for either anomaly or the background.

\subsection{Overfitting Prevention and Early Stop}
We use a small mini-batch size (32) in training, and use weight decay to reduce the possibility of overfitting. The extensive data augmentation during training helps the model generalize and train as completely as possible. We also validate the network's performance every iteration using the testing dataset and stop the training early.

\section{Testing and Experimental Results}
There are only 69628 parameters in our weight file (283KB in size) and our network is able to process more than 1000 RxMER captures in one second on a single CPU core. During testing, we use mAP as the metric to indicate the performance of the model. The confidence score threshold is 0.5 and NMS threshold is 0.5. NMS is performed on a per-class basis using IoU thresholds 0.5 and 0.75. We also performed testing with soft-NMS\cite{soft_nms} with very minor code changes, and slightly improved the model's mAP. The inference performance is compared to the results from a previous research paper\cite{access_network_scte}.
\begin{table}[h!]
\renewcommand{\arraystretch}{1.3}
\caption{Mean average precision (\%) per class}
\label{tab:map_per_class}
\centering
\begin{tabular}{|p{1.3cm}|p{0.7cm}p{0.7cm}p{0.7cm}|p{0.7cm}p{0.7cm}p{0.7cm}|}
    \hline
     & $mAP_{50}$ & $mAP_{50}$ (soft-NMS) & $mAP_{50}$ (Previous) & $mAP_{75}$ & $mAP_{75}$ (soft-NMS) & $mAP_{75}$ (Previous)\\
    \hline
    
    \hline

    LTE ingress & 99.21 & \textbf{99.23} & 73.67 & 99.13 & \textbf{99.14} & 60.09\\
    \hline

    Suck-out & 98.35 & \textbf{98.44} & 62.18 & 96.16 & \textbf{96.25} & 48.39\\
    \hline
    
    Wave & 98.42 & 98.42 & 58.93 & 98.42 & 98.42 & 56.82\\
    \hline
    
    Roll-off & 99.15 & 99.15 & 87.36 & 97.26 & 97.26 & 70.61\\
    \hline
    
    Spike & 93.64 & \textbf{93.89} & 68.73 & 93.57 & \textbf{93.83} & 59.46\\
    \hline
\end{tabular}
\end{table}
\begin{table}[h!]
\renewcommand{\arraystretch}{1.3}
\caption{Overall mean average precision (\%)}
\label{tab:map_all}
\centering
\begin{tabular}{|p{0.7cm}p{0.7cm}p{0.7cm}|p{0.7cm}p{0.7cm}p{0.7cm}|}
    \hline
    $mAP_{50}$ & $mAP_{50}$ (soft-NMS) & $mAP_{50}$ (Previous) & $mAP_{75}$ & $mAP_{75}$ (soft-NMS) & $mAP_{75}$ (Previous)\\
    \hline
    
    \hline

    97.75 & \textbf{97.82} & 70.17 & 96.91 & \textbf{96.98} & 59.07\\
    \hline
\end{tabular}
\end{table}

From the results in Table \ref{tab:map_per_class} and Table \ref{tab:map_all} we can see that the algorithm we propose in this paper outperforms the algorithm from the previous research by a large margin, part of the reason is that the previous algorithm is not designed to differentiate all the 5 types of anomalies and it has non-ideal region proposal generation. The previous algorithm also predicts much slower at about 50 samples per second. On the other hand, Soft-NMS improves mAP on small and medium sized anomalies in our evaluation. The small difference between mAP$_{50}$ and mAP$_{75}$ indicates that our network produces high quality localization prediction. The lower mAP$_{50}$ and mAP$_{75}$ on spike detection are caused by lower recall which indicates that our network has difficulties recognizing all of the smallest anomalies especially when they are densely located. This can be caused by that the prediction layers have at most 52 grids in prediction resolution. One grid can produce 3 predictions in which only 1 of them is trained to recognize the anomaly, which means small anomalies that are close to each other can be missed by the detector. The performance of our network on small and dense anomalies can possibly be improved by introducing an additional prediction layer with 104 grids or larger input sizes such as 608 or 928. However, in our use case, spikes on OFDM RxMER captures are minor issues that do not warrant attention by repair technicians or network operations center personnel. Therefore, it is not necessary to trade inference performance for spike detection improvement.

\begin{figure}[h!]
  \centering
  \includegraphics[width=0.37\textwidth]{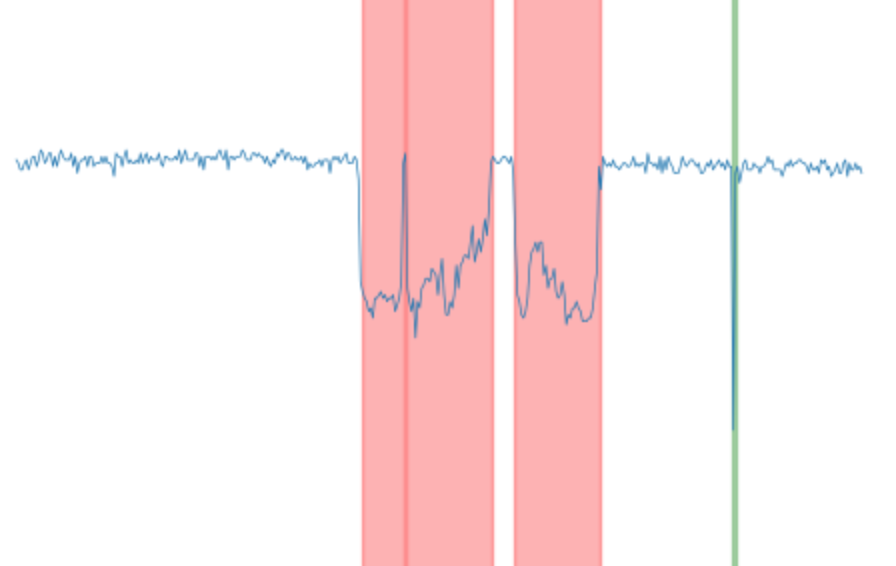}
  \caption{Detection result: multiple LTE ingress points and a spike}
  \label{lte_and_spikes_more}
\end{figure}

\begin{figure}[h!]
  \centering
  \includegraphics[width=0.37\textwidth]{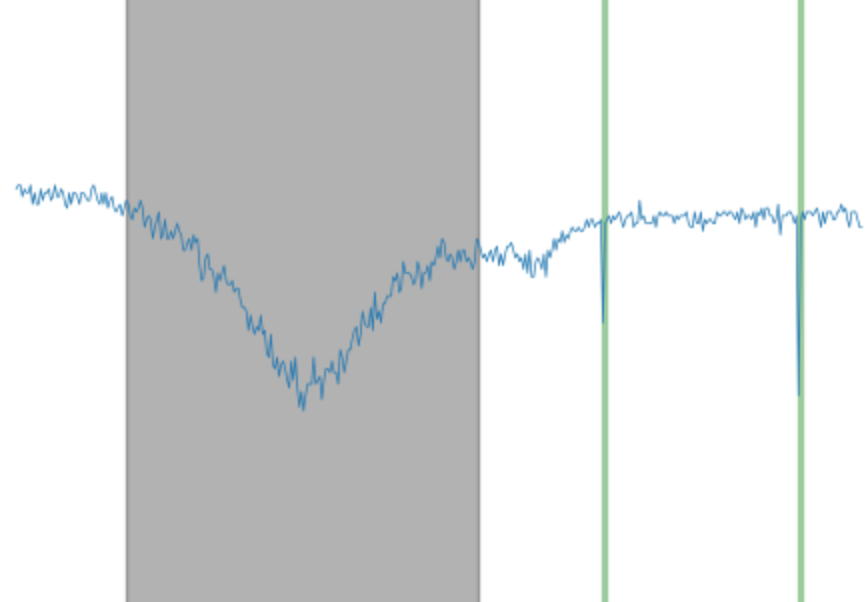}
  \caption{Detection result: a suck-out and 2 spikes}
  \label{suck_out_and_spike}
\end{figure}

\begin{figure}[h!]
  \centering
  \includegraphics[width=0.37\textwidth]{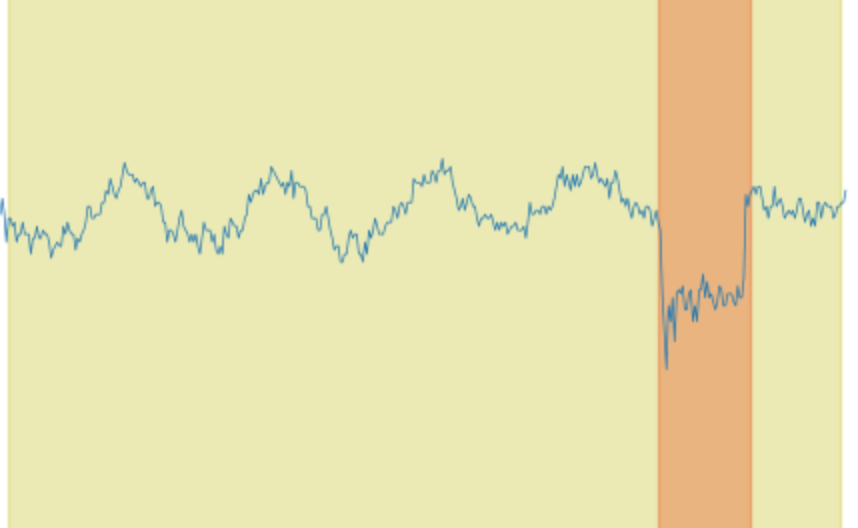}
  \caption{Detection result: wave and a LTE ingress point}
  \label{wave_and_lte}
\end{figure}

\begin{figure}[h!]
  \centering
  \includegraphics[width=0.37\textwidth]{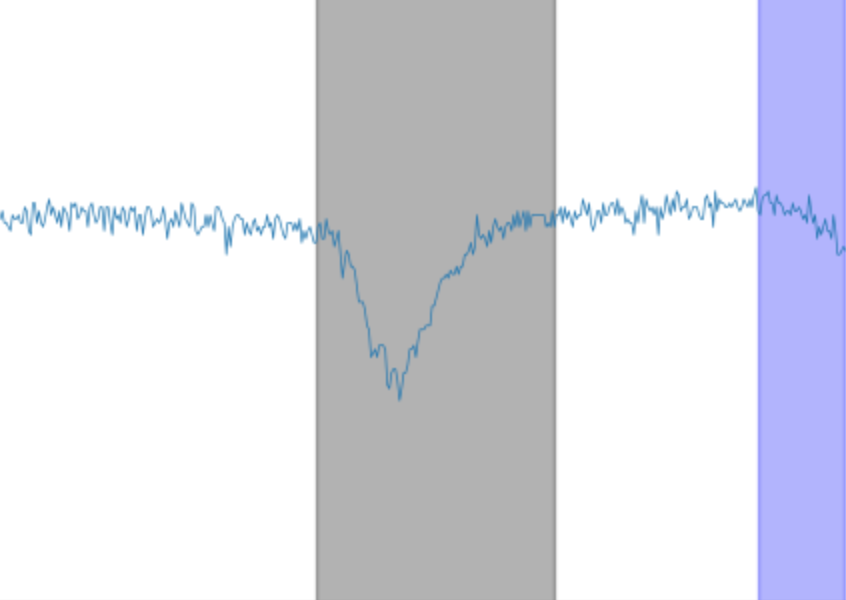}
  \caption{Detection result: a suck-out and right roll-off}
  \label{suck_out_and_roll_off}
\end{figure}

\begin{figure}[h!]
  \centering
  \includegraphics[width=0.37\textwidth]{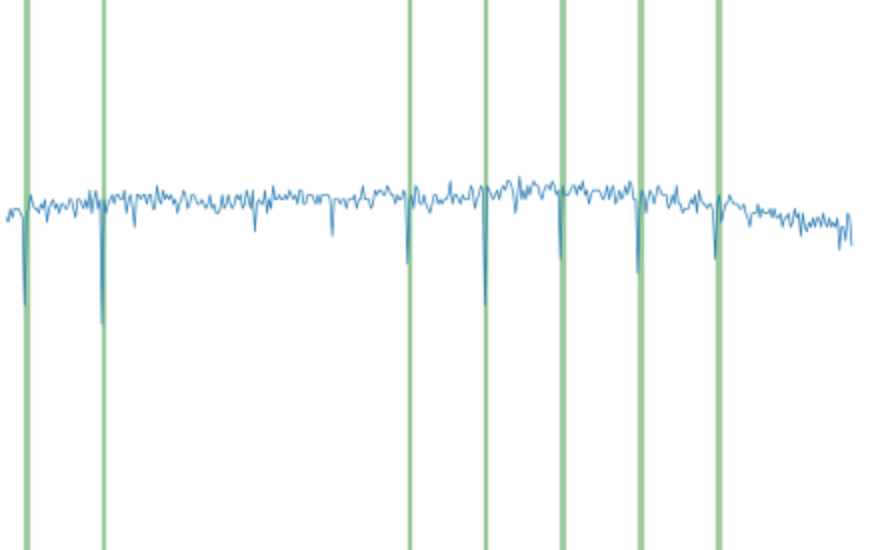}
  \caption{Detection result: many spikes across the spectrum}
  \label{many_spikes}
\end{figure}

\begin{figure}[h!]
  \centering
  \includegraphics[width=0.37\textwidth]{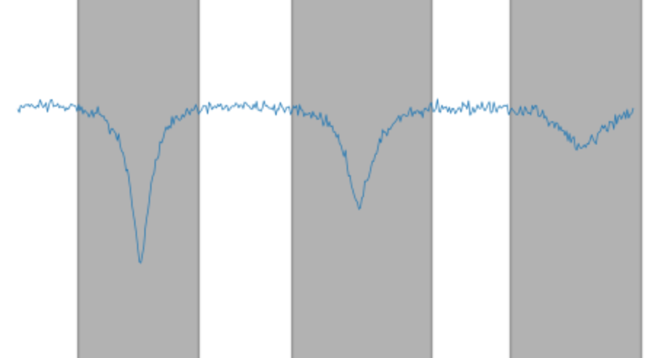}
  \caption{Detection result: many suck-outs across the spectrum}
  \label{many_suckouts}
\end{figure}

\begin{figure}[h!]
  \centering
  \includegraphics[width=0.37\textwidth]{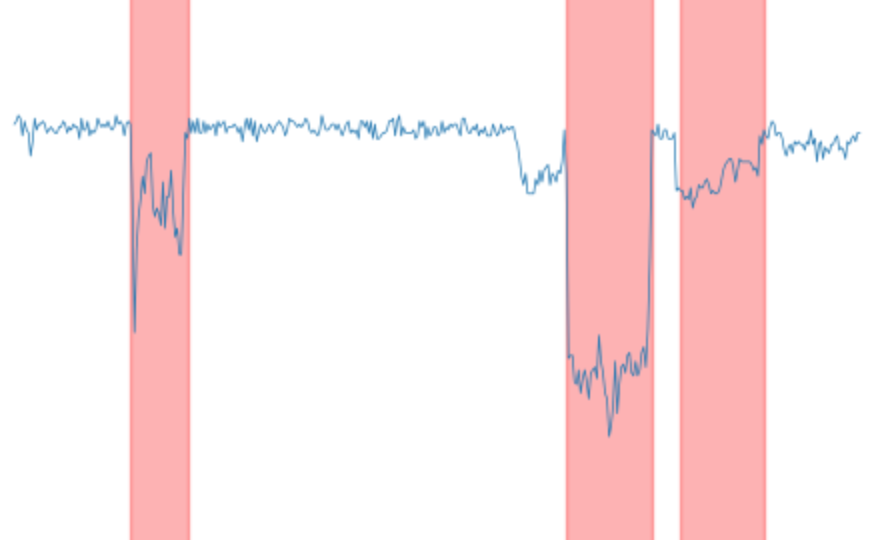}
  \caption{Detection result: a small LTE impairment is not detected}
  \label{missing_small_lte}
\end{figure}

\begin{figure}[h!]
  \centering
  \includegraphics[width=0.37\textwidth]{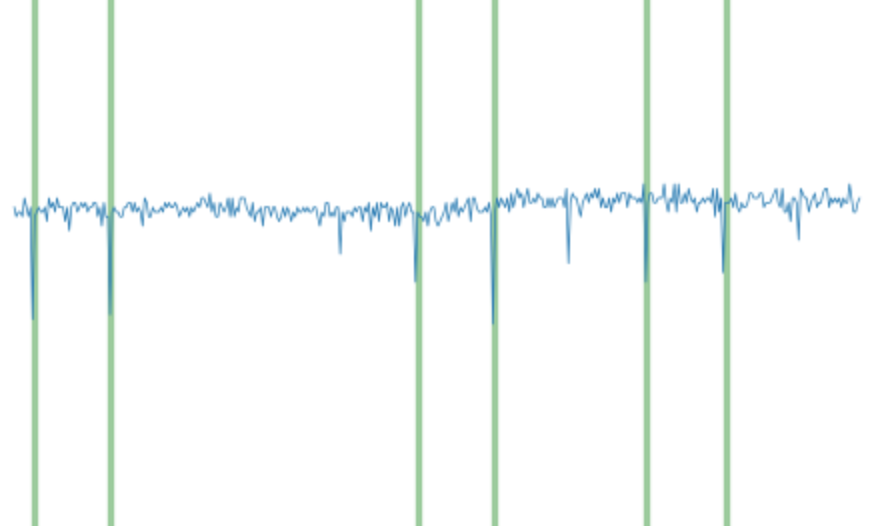}
  \caption{Detection result: a few spike impairments are not detected}
  \label{missing_spikes}
\end{figure}

\begin{figure}[h!]
  \centering
  \includegraphics[width=0.37\textwidth]{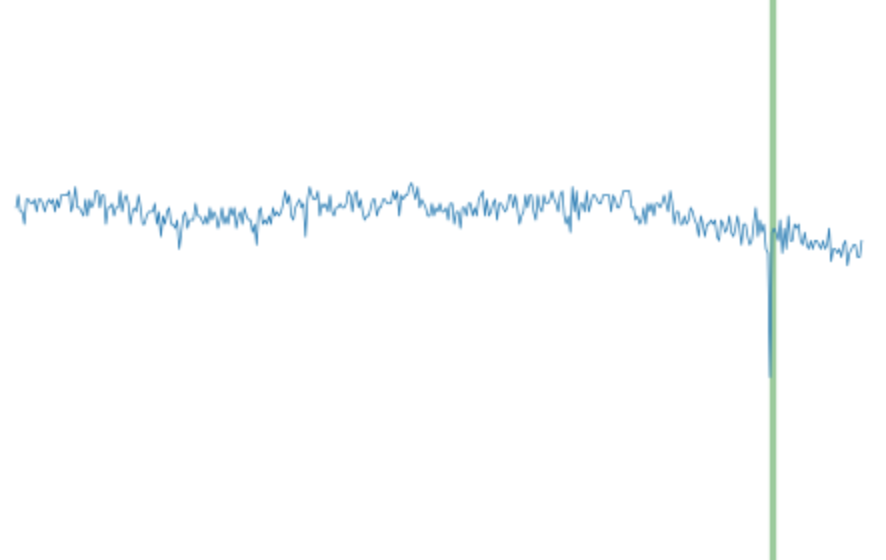}
  \caption{Detection result: a subtle wave is not detected}
  \label{missing_wave}
\end{figure}

\begin{figure}[h!]
  \centering
  \includegraphics[width=0.37\textwidth]{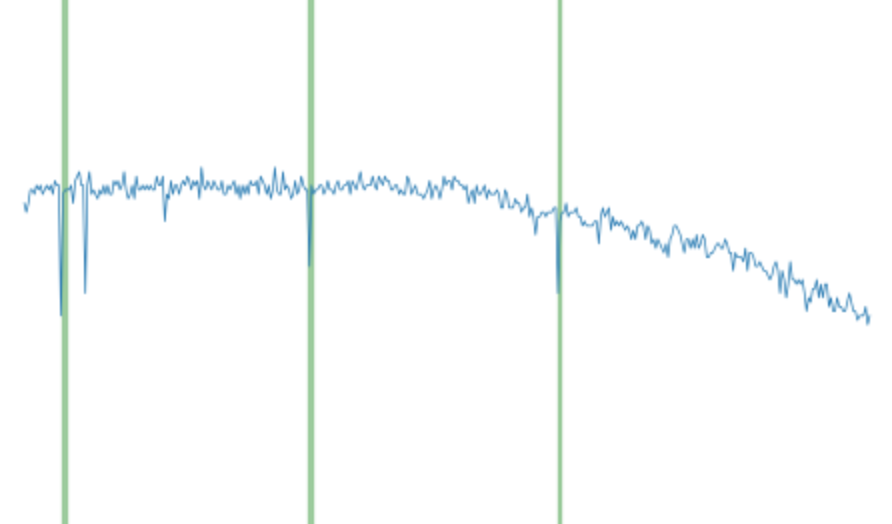}
  \caption{Detection result: a roll-off on half of the OFDM channel is not detected}
  \label{missing_roll_off}
\end{figure}

We list example detection results from Fig.\ref{lte_and_spikes_more} to Fig.\ref{missing_roll_off}. Fig.\ref{lte_and_spikes_more} shows 3 LTE channels that are interfering with the modem's OFDM channel, there is also a spike on the higher frequency end; Fig.\ref{suck_out_and_spike} shows a large suck-out on the lower frequency end of the spectrum and 2 spikes; Fig.\ref{wave_and_lte} shows a wave that is interfering the whole channel and a LTE channel that is affecting the higher frequency end; Fig.\ref{suck_out_and_roll_off} shows a suck-out in the middle of the OFDM channel and a right roll-off; Fig.\ref{many_spikes} shows 7 spikes detected across the channel; Fig.\ref{many_suckouts} shows a rare capture that 3 suck-outs exist on 1 OFDM channel; Fig.\ref{missing_small_lte} shows that a small LTE impairment is not detected, which could be caused by that the small LTE impairment is in the same prediction grid with the larger LTE impairment on its right side, and one prediction grid can only detect one anomaly at a time; Fig.\ref{missing_spikes} shows that some small spikes are not detected, which could be caused by that after many layers of feature extraction, some details are lost; Fig.\ref{missing_wave} shows that a very subtle wave across the OFDM channel is not detected; Fig.\ref{missing_roll_off} shows that a large roll-off is not detected.

\section{Conclusion and Future Work}
Our key contributions in this paper can be summarized as follows
\begin{itemize}
\item \textit{A novel anomaly detector implementation based on YOLOv3's network architecture}: Based on YOLOv3's network architecture, we develop a Fully 1D-Convolutional Neural Network with 45 1D-Convolutional layers for feature extraction with significantly less filters. Shortcut layers and route layers\cite{yolov2} are used with 3 prediction layers to aggregate features from different scales and perform prediction at different scales. This anomaly detector achieves very low resource consumption, very high performance, and very high mAP. In our evaluation using cable modems' downstream RxMER data, the anomaly detector processes more than 1000 cable modems' data captures every second using a single CPU core at up to 600 KB memory consumption. Each cable modems' data capture is a 1-D array that has 1800 to 2000 MER values.

\item \textit{A specifically designed down-sampling algorithm which significantly improves performance while keeping important features}: The down-sampling algorithm is specifically designed to pre-process 1-dimensional input data for performance gains. It significantly reduces the number of values from the original input and keeps important features for anomaly detection.

\item \textit{Data augmentation techniques for anomaly detection on data series}: We specifically develop data augmentation techniques to improve the training quality, model generalization, and number of training and testing samples. These techniques significantly reduce the amount of manual work and improve the model's performance.

\item \textit{Demonstration of leveraging state-of-the-art object detection algorithm in 1-dimensional anomaly detection problems}: We successfully leverage the state-of-the-art object detection algorithm and transform it into a promising problem solution for anomaly detection by changing its network architecture, the loss function, data augmentation techniques, and significantly improving its performance.
\end{itemize}

PNM is a field of work in the cable industry that is constantly extending with different architectures and updates to protocols. As ProOps suggests a structure that we believe can be a model for the industry, built on a model for data collection for the industry as well, we intend to continue research, prototyping, and validating new PNM solutions for new updates to DOCSIS networking and technologies involving optical networks\cite{coherent_scte} as well.

The anomaly detection solution reported here is very flexible, applicable to a wide range of related problems, and is easily extensible. For medium and large impairments, which are the most important to address, the mAP of our solution is over 98\%. Because our model has superb inference performance, we can focus on improvements in three fronts: additional measures that may provide additional evidence to improve precision, focusing on the most severe issues that are more likely to be problems that need to be addressed, and collecting RxMER data from neighboring CMs or the same CM over time to confirm the found issues. As we extend the solution to other data sources and over time series, we expect mAP to be higher. Using ProOps as the platform allows us to implement this approach and adjust settings rapidly as well. But we are already at a level of performance that a network operations center person can quickly confirm the most severe problems and gain order of magnitude efficiencies with human time and expertise. We also intend to extend the anomaly detection engine into other spectrum and time series measures, including spectrum analysis data, pre-equalization coefficients data, various time series statistics for specific CMs, and perhaps even clustering of impairments over time and over CMs to improve repair efficiency.

To further improve our detector's performance, there are other state-of-the-art object detection networks we can learn from such as EfficientDet\cite{efficientdet} which uses a weighted bi-directional feature pyramid network (BiFPN) and a compound scaling method to reduce network size, improve inference performance, and improve mAP. The prediction grid layout can be designed to improve performance on small and dense anomalies. We can also experiment with neural network based NMS\cite{learnnms}.

\section*{Acknowledgment}
We would like to thank CableLabs members who provided the data that helped with this research, and CableLabs for supporting the work.

\ifCLASSOPTIONcaptionsoff
  \newpage
\fi

\end{document}